%% file: main.tex
\documentclass[10pt,twocolumn,letterpaper]{article}

\usepackage{cvpr}              
\usepackage{multirow}
\usepackage{booktabs}
\usepackage{pifont}
\usepackage{colortbl}
\usepackage{algorithm}
\usepackage{algorithmic}
\usepackage[export]{adjustbox}

\definecolor{my_green}{RGB}{51,102,0}
\definecolor{my_red}{RGB}{204, 0, 0}

\definecolor{paired-light-blue}{RGB}{198, 219, 239}
\definecolor{paired-dark-blue}{RGB}{49, 130, 188}
\definecolor{paired-light-orange}{RGB}{251, 208, 162}
\definecolor{paired-dark-orange}{RGB}{230, 85, 12}
\definecolor{paired-light-green}{RGB}{199, 233, 193}
\definecolor{paired-dark-green}{RGB}{49, 163, 83}
\definecolor{paired-light-purple}{RGB}{218, 218, 235}
\definecolor{paired-dark-purple}{RGB}{117, 107, 176}
\definecolor{paired-light-gray}{RGB}{217, 217, 217}
\definecolor{paired-dark-gray}{RGB}{99, 99, 99}
\definecolor{paired-light-pink}{RGB}{222, 158, 214}
\definecolor{paired-dark-pink}{RGB}{123, 65, 115}
\definecolor{paired-light-red}{RGB}{231, 150, 156}
\definecolor{paired-dark-red}{RGB}{131, 60, 56}
\definecolor{paired-light-yellow}{RGB}{231, 204, 149}
\definecolor{paired-dark-yellow}{RGB}{141, 109, 49}  
\definecolor{myblue}{RGB}{218,232,252}
\definecolor{mygray}{RGB}{220,220,220}
\definecolor{mypink}{RGB}{251,49,153}

\definecolor{cvprblue}{rgb}{0.21,0.49,0.74}
\usepackage[pagebackref,breaklinks,colorlinks,allcolors=cvprblue]{hyperref}

\def\confYear{2025}

\title{LeanVAE: An Ultra-Efficient Reconstruction VAE for Video Diffusion Models}

\author{
    Yu Cheng, Fajie Yuan \\  
    Westlake University \\
    {\tt\small \{chengyu, yuanfajie\}@westlake.edu.cn}
}

\begin{document}

\maketitle

\input{sec/0_abstract}    
\input{sec/1_intro}

\input{sec/5_related_works}

\input{sec/2_preliminary}

\input{sec/3_method}

\input{sec/4_experiments}

\input{sec/6_conclusion}

\newpage

{
    \small
    \bibliographystyle{ieeenat_fullname}
    \bibliography{main}
}

\input{sec/X_suppl}

\end{document}

%% file: sec/0_abstract.tex
\begin{abstract}
Recent advances in Latent Video Diffusion Models (LVDMs) have revolutionized video generation by leveraging Video Variational Autoencoders (Video VAEs) to compress intricate video data into a compact latent space. However, as LVDM training scales, the computational overhead of Video VAEs becomes a critical bottleneck, particularly for encoding high-resolution videos. To address this, we propose \textbf{LeanVAE}, a novel and ultra-efficient Video VAE framework that introduces two key innovations: (1) a lightweight architecture based on a Neighborhood-Aware Feedforward (NAF) module and non-overlapping patch operations, drastically reducing computational cost, and (2) the integration of wavelet transforms and compressed sensing techniques to enhance reconstruction quality. Extensive experiments validate LeanVAE’s superiority in video reconstruction and generation, particularly in enhancing efficiency over existing Video VAEs.  Our model offers up to 50× fewer FLOPs and 44× faster inference speed \footnote{Evaluated on $768 \times 768$ videos.} while maintaining competitive reconstruction quality, providing insights for scalable, efficient video generation. Our models and code are available at \href{https://github.com/westlake-repl/LeanVAE}{https://github.com/westlake-repl/LeanVAE}.
\end{abstract}

\vspace{-1.2em}

%% file: sec/1_intro.tex
\section{Introduction}
\label{sec:intro}

Recent advances in video generation models have garnered widespread attention due to their significant impact on fields such as animation, advertisement, and simulation~\cite{videoworldsimulators2024}. Among these, Latent Video Diffusion Models (LVDMs) have emerged as a leading approach. In these models, Video Variational Autoencoders (Video VAEs) facilitate a bidirectional mapping between the high-dimensional, redundant video pixel space and a low-dimensional, compact latent space, enabling both spatial and temporal compression of the 3D RGB volume. A diffusion denoising model is then trained to model the distribution of the latent space. This latent space design offers substantial advantages in training efficiency, stability, and scalability compared to pixel-space diffusion models, leading to the widespread adoption of LVDMs in notable works such as Open-Sora-Plan~\cite{pku_yuan_lab_and_tuzhan_ai_etc_2024_10948109}, Open-Sora~\cite{opensora}, Latte~\cite{ma2024latte}, CogVideoX~\cite{yang2024cogvideox}, EasyAnimate~\cite{xu2024easyanimate}, HunyuanVideo~\cite{kong2024hunyuanvideo}, and Cosmos~\cite{agarwal2025cosmos}.

The Video VAEs play a crucial role in LVDM models. The quality of the video reconstruction by the VAE directly determines the upper bound of the generated video quality in LVDMs. Artifacts~\cite{1530348} such as blurring, distortions, and motion inconsistencies in VAE outputs directly degrade the final generation. Besides, the encoding efficiency of the VAE often becomes a computational bottleneck during LVDM training~\cite{li2024wf}, particularly for long-duration, high-resolution videos. Therefore, optimizing the Video VAE is essential to enable efficient large-scale pre-training for LVDMs.

Given the significant value of an efficient and powerful Video VAE in LVDMs, many research has been devoted to this area. Early works like OD-VAE~\cite{chen2024od}, CV-VAE~\cite{zhao2024cv}, and SliceVAE~\cite{xu2024easyanimate} typically extended the well-known Stable Diffusion image VAE (SD VAE)~\cite{rombach2022high} by inflating 2D convolutions to 3D and incorporating spatiotemporal attention modules to achieve temporal and spatial compression for videos. While these models demonstrated promising results, they incur prohibitive computational costs. For example, OD-VAE consumes around 32GB of memory to encode and decode a 5-frame 1080 square video with FP16 inference. Subsequent research has pursued two main strategies to address the computational burden. In works like Open-Sora, Movie Gen~\cite{polyak2024moviegencastmedia}, Cosmos Tokenizer~\cite{agarwal2025cosmos}, and VidTok~\cite{tang2024vidtok}, various techniques have been proposed to factorize the computationally intensive 3D structures. Other works focus on building simplified network architectures. For example, Cosmos Tokenizer~\cite{agarwal2025cosmos}, WF-VAE~\cite{li2024wf} reduce network redundancies by introducing wavelet transforms. Omnitokenizer~\cite{wang2024omnitokenizer}, ViTok~\cite{hansen2025learnings} build much smaller transformer-based models by leveraging global attention, though they suffer from quadratic computational complexity with respect to video resolution. 

\begin{figure*}[ht]
	\centering    \includegraphics[width=\linewidth]{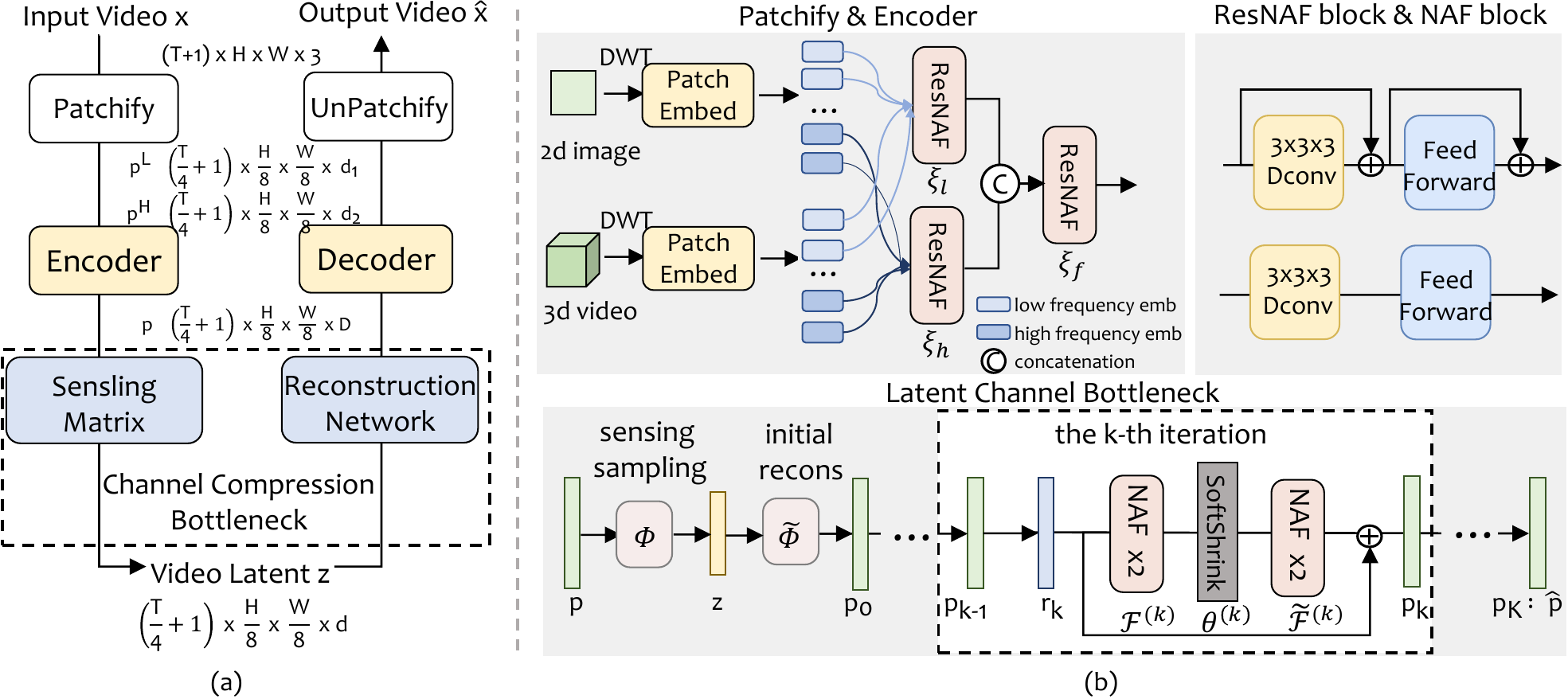}
	\vspace{-2.0em}
	\caption{(a) LeanVAE framework overview. (b) Key components: \textbf{Patchifier} for image-video joint patching in frequency domain; \textbf{\textit{Encoder}} for hierarchical feature extraction; \textbf{(Res)NAF} serves as model backbone, enabling Neighborhood-Aware Feedforward (with Residual connections); \textbf{Latent Channel Bottleneck} for latent channel compression and restoration based on $\textit{ISTA-Net}^{+}$ algorithm.}

	\label{fig:main}
	\vspace{-1.2em}
\end{figure*}

Inspired by above works, we aim to advance Video VAE and demonstrate a significant improvement in balancing efficiency and performance by proposing a novel and highly efficient Video VAE approach, coined LeanVAE. Unlike previous methods that either inherited the Video VAE from the SD VAE or employed ViT~\cite{arnab2021vivit} architectures, we use two key designs to ensure a lightweight network. First, we employ a non-overlapping patch operation to downsample both spatial and temporal dimensions at the very beginning of the network, drastically reducing the computation. Second, we introduce a Neighborhood-Aware Feedforward (NAF) module as the backbone of our model, as shown in ~\cref{fig:main}(b). The NAF module efficiently captures local contexts via depthwise convolutions and processes them through a feedforward network, enabling lightweight yet powerful extraction and transformation of video features.

After that, we integrate wavelet transform~\cite{burrus1998wavelets} and Compressed Sensing (CS)~\cite{jeong2024compressed} technique to further enhance reconstruction quality. Specifically, we apply wavelet transforms to the input RGB signal following prior work~\cite{li2024wf}, obtaining features in the frequency domain that enrich the input space and aid in expanding the model's receptive field~\cite{kim2024hybrid}. Another key improvement is the adoption of a classical CS algorithm for latent channel dimension compression. This addresses the inefficiency we identified in conventional two-linear-layer AutoEncoding (AE) ~\cite{jeong2024compressed}  structure for this task.
 To the best of our knowledge, this is the first application of CS for channel compression bottleneck in Video VAEs. Our experiments demonstrate a significant performance boost, revealing the immense potential of CS in this domain.

Finally, we explore the design space of LeanVAE and propose several variations for further optimization. Through careful experimentation, we identify the optimal configuration and train a series of highly efficient LeanVAE models. We then demonstrate its superiority in video reconstruction against various strong baselines and highlight its strengths in both video reconstruction and generation. \cref{fig:SpeedMemoryTest} illustrates its ability to achieve high reconstruction quality while significantly reducing computational cost, with this advantage especially pronounced at higher resolutions. It is also worth mentioning that LeanVAE supports joint modeling of images and videos and maintains causal property of the latent space, making it broadly applicable. Additionally, we identify that the common practice in ViT of applying LayerNorm~\cite {ba2016layer} to pixel patches is a key factor contributing to block artifacts~\cite{li2022variable} in reconstruction. We believe this is an interesting discovery that may inspire improvements in similar models for low-level vision tasks.

In summary, we present a highly efficient and lightweight video reconstruction framework that accelerates Video VAE and improves training throughput for LVDMs. We believe this approach marks a important enhancement toward more scalable and efficient video generation. There are three technical novelty that contributed to the good performance and merits of LeanVAE: 
\begin{itemize}
\item	We adopt the patch method inspired by ViT and design a lightweight NAF module as the model backbone, achieving a ultra-efficient network architecture. 
\item	We enrich input representations using wavelet transforms and, for the first time, introduce a CS framework as the channel compression bottleneck in Video VAE. Both techniques significantly enhance model performance. 
\item	We conduct an in-depth architectural ablation for further optimization. Extensive experiments validate the superiority of LeanVAE in both video reconstruction and generation, particularly in enhancing efficiency over existing Video VAEs. 
\end{itemize}

%% file: sec/5_related_works.tex
\section{Related Work}
\label{sec:related_work}
VAEs are powerful models for learning high-dimensional data distributions in either continuous or discrete~\cite{villegas2022phenaki,ge2022long, yu2024languagemodelbeatsdiffusion} latent space. This work focuses on continuous VAEs, which are widely used in latent diffusion models to reduce the computational cost. We categorize related work into two main directions: Regular Video VAEs and Specific Video VAEs.

\noindent\textbf{Regular Video VAEs }\quad
Regular Video VAEs compress the time, height, and width dimensions of videos into 3D latent representations that resemble visual thumbnails. These models typically achieve a 4×8×8 compression ratio for the original videos and are widely adopted in popular open-source video generation frameworks~\cite{opensora, pku_yuan_lab_and_tuzhan_ai_etc_2024_10948109,  yang2024cogvideox, xu2024easyanimate, agarwal2025cosmos}. To construct a Video VAE, the most straightforward idea is to inherit from the well-established SD image VAE to leverage its spatial compression priors. This involves inflating 2D convolutions in the image model to 3D convolutions as the video counterpart. Building on this, various techniques have been introduced to improve reconstruction quality. Pioneering works like OD-VAE~\cite{chen2024od}, CV-VAE~\cite{zhao2024cv}, and SliceVAE~\cite{xu2024easyanimate} introduced attention layers to model global spatiotemporal dependencies. Models like IV-VAE~\cite{wu2024improved}, VideoVAE+~\cite{xing2024large}, and FILM-VAE~\cite{argawhigh} proposed well-curated spatiotemporal convolutional modules for smoother motion and enhanced spatial details. CogVideoX~\cite{yang2024cogvideox} introduced increasing the latent channel dimension for better reconstruction. These innovations have effectively pushed the boundaries of reconstruction quality. 

However, as video diffusion models scale, the computational bottleneck caused by Video VAE models in the LVDM pipeline has become increasingly unaffordable, leading to growing interest in a better balance between efficiency and reconstruction performance. Some works have attempted to decompose the 3D network structure required for spatiotemporal compression to reduce computation. In Open-Sora~\cite{opensora} and MovieGen~\cite{polyak2024moviegencastmedia}, 2D+1D network architectures are chosen to compress the spatial dimensions first, followed by temporal compression. Cosmos Tokenizer~\cite{agarwal2025cosmos} replaced standard 3D convolutions with factorized convolutions, applying 2D spatial convolutions followed by a temporal convolution. VidTok~\cite{tang2024vidtok} further incorporates AlphaBlender into this factorized structure for better temporal modeling. Other works focus on approaches to simplify the network architecture. In Cosmos Tokenizer~\cite{agarwal2025cosmos} and WF-VAE~\cite{li2024wf}, wavelet transforms are applied to videos, resulting in lighter VAE models that reduce computation while maintaining good reconstruction quality. Transformer-based methods like OmniTokenizer~\cite{wang2024omnitokenizer}, and ViTok~\cite{hansen2025learnings} leveraged global dependency modeling to build models with much fewer parameters, but they suffer quadratic complexity with respect to video resolution.  Building on these insights, our work achieves an excellent balance between efficiency and quality by cleverly applying wavelet transforms and compressive sensing algorithms, along with a carefully designed lightweight network architecture.

\noindent\textbf{Specific Video VAEs }\quad
In LVDM, some works focus on training VAEs to generate specialized video latent spaces that enable more efficient training of diffusion models. For instance, HVDM~\cite{kim2024hybrid} uses a hybrid video autoencoder to obtain a latent space that combines 2D triplane representations with 3D volume information. VidTwin~\cite{wang2024vidtwin} incorporates carefully designed submodules to disentangle the structure and dynamics in the latent space. LTX-Video~\cite{hacohen2024ltx} proposes a video compression method that increases spatiotemporal compression ratios and latent channel dimension simultaneously so as to relocate the patchifying operation from the diffusion model to the VAEs. These works require designing corresponding diffusion models to accommodate the unique characteristics of the latent space. Due to differences in compression ratios and latent space properties compared to regular Video VAEs, we consider these works as future directions for research rather than baselines for comparison.

%% file: sec/2_preliminary.tex
\section{Preliminary}
In this section, we briefly introduce two key techniques utilized in LeanVAE: Wavelet Transform and Compressed Sensing.
\subsection{Wavelet Transform}
Wavelet transform is widely used in fields such as medical imaging and video compression. It decomposes visual signals in the RGB space into multiple frequency-domain components.  In our work, we use the Haar wavelet transform to process images and videos, which includes a pair of reversible processes: Discrete Wavelet Transform (DWT) and  Inverse Discrete Wavelet Transform (IDWT).  
For a 2D image $\mathbf{I} \in \mathbb{R}^{c \times h \times w}$, where $c$, $h$, and $w$ represent channels, height, and width, respectively, the Haar 2D kernels operate along the height and width axes, decomposing the image into four subbands: $\{ \mathbf{I}_{ll}, \mathbf{I}_{lh}, \mathbf{I}_{hl}, \mathbf{I}_{hh} \}$, each of size $c \times \frac{h}{2} \times \frac{w}{2}$. Similarly, given a 3D video $\mathbf{V} \in \mathbb{R}^{c \times t \times h \times w}$,  where $t$ denotes the temporal frames, the 3D Haar DWT is performed, dividing the input volume into eight subbands: $\{ \mathbf{V}_{lll}, \mathbf{V}_{llh}, \mathbf{V}_{lhl}, \mathbf{V}_{lhh}, \mathbf{V}_{hll}, \mathbf{V}_{hlh}, \mathbf{V}_{hhl}, \mathbf{V}_{hhh} \}$, each of size $c \times \frac{t}{2} \times \frac{h}{2} \times \frac{w}{2}$.  
Among these subbands, the first component $(\mathbf{I}_{ll} or \mathbf{V}_{lll}) $ represents low-frequency components providing long-term average information. The remaining components capture high-frequency details and local changes. These subbands can be perfectly reconstructed into the original signal using IDWT, as presented in the Appendix ~\cref{fig:wavelet}. After applying the Haar DWT, the spatiotemporal dimensions of the input visual signal are halved, while the channel dimension increases accordingly. This enriches the feature space and has been shown to effectively expand the receptive field of CNNs~\cite{kim2024hybrid}.

\subsection{Compressed Sensing}
Compressed sensing is a promising technique that enables signal reconstruction with far fewer measurement data than required by the Nyquist-Shannon sampling theorem~\cite{zhang2018ista}. Over the past decade, numerous effective CS methods have been developed for signal compression and restoration, which expresses the original signal with a reduced number of measurements and restores the signal through a recovery algorithm. For a signal $\mathbf{p} \in \mathbb{R}^D$, sensing matrix $\mathbf{\Phi} \in \mathbb{R}^{d \times D}$, the measurements $\mathbf{z} \in \mathbb{R}^d (d \ll D)$ is obtained by the sampling process:
\[
\mathbf{z} = \mathbf{\Phi}\mathbf{p}.
\]
To achieve reliable reconstruction $\hat{\mathbf{p}}$ from reduced measurements $\mathbf{z}$, CS methods typically solve the following optimization problem: 
\begin{equation}
	\mathop{\mathbf{argmin}}\limits_{\mathbf{p}}{\;} \frac{1}{2}\Vert{\mathbf{z} - \mathbf \Phi \mathbf{p}}\Vert_2^2 + \lambda\Vert{\mathbf{\Psi} 
		\mathbf{p}}\Vert_1,
	\label{Ax-y}
\end{equation}
where $\Vert{\mathbf{z} - \mathbf \Phi \mathbf{p}}\Vert_2^2$ ensures data fidelity and $\Vert{\mathbf{\Psi} \mathbf{p}}\Vert_1$ promotes sparsity in $\mathbf{\Psi} \mathbf{p}$, with $\lambda$ as the regularization parameter. 

In this work, we use CS algorithm to build the latent channel bottleneck, which performs downsampling and reconstruction of the latent channels. To be specific, we employ $\textit{ISTA-Net}^{+}$ ~\cite{zhang2018ista}, a unfolding algorithm framework that effectively optimizes ~\cref{Ax-y} through K iterative steps. In  the $k$-\textit{th} iteration, it sequentially updates the intermediate reconstruction result $\mathbf{r}^{{(k)}}$ and final reconstruction $\mathbf{p}^{{(k)}}$ as follows:

\begin{equation} \label{eq: update r module}
	\setlength{\abovedisplayskip}{5pt}
	\setlength{\belowdisplayskip}{5pt}
	\mathbf{r}^{(k)} = \mathbf{p}^{(k-1)} - \rho^{(k)} \mathbf{\Phi}^{\top} (\mathbf{\Phi} \mathbf{p}^{(k-1)} - \mathbf{z}).
\end{equation}
\begin{equation} \label{eq: update x module}
	\setlength{\abovedisplayskip}{5pt}
	\setlength{\belowdisplayskip}{5pt}
	\mathbf{p}^{(k)} = \mathbf{r}^{(k)} +\widetilde{\mathcal{F}}^{(k)}(soft(\mathcal{F}^{(k)}(\mathbf{r}^{(k)}), \theta^{(k)} )).
\end{equation}

The reconstruction network is parameterized by  
\(\mathbf{\Theta} = \{\rho^{(k)}, \theta^{(k)}, \mathcal{F}^{(k)}, \widetilde{\mathcal{F}}^{(k)}\}_{k=1}^{K}\),  
where \(\mathcal{F}^{(k)}\) and \(\widetilde{\mathcal{F}}^{(k)}\) represent the forward and backward networks at the \(k\)-th iteration, respectively, and \(\rho^{(k)}\) and \(\theta^{(k)}\) are scalar parameters. The term \(soft\) refers to the soft-shrinkage threshold operation:
\begin{equation}
	\label{soft}
	soft(\boldsymbol{x}, \theta) = 
	\text{sgn}(\boldsymbol{x}) \cdot \max\left(|\boldsymbol{x}| - \theta, 0\right)
\end{equation}

%% file: sec/3_method.tex
\section{Method}
\subsection{Overall Process}
Let the input video be denoted  as $\mathbf{x} \in \mathbb{R}^{(T+1) \times H \times W \times 3}$,  where (1 + T) is the number of frames(with T = 0 for images) and H × W denotes the spatial resolution. As shown on the left side of ~\cref{fig:main}(a), the input video $\mathbf{x}$ is encoded into the compressed latent $\mathbf{z} \in \mathbb{R}^{(T^{'}+ 1) \times H^{'} \times W^{'} \times d}$,with a spatial compression factor of $c_t = \frac{T}{T^{'}}$, a temporal compression factor of $c_s = \frac{H}{H^{'}} = \frac{W}{W^{'}}$, and d is the latent channel dimension.  The video is first processed with a patchify operation to obtain patch embeddings, including high-frequency component (HC) embedding $\mathbf{p^{H}} \in \mathbb{R}^{(T^{'}+1) \times H^{'} \times W^{'} \times d_{1}}$ and low-frequency component (LC) embedding $\mathbf{p^{L}} \in \mathbb{R}^{(T^{'}+1) \times H^{'} \times W^{'} \times d_{2}}$. These embeddings are then passed through the \textit{Encoder} to facilitate feature interaction and transformation, producing the feature tensor $\mathbf{p}\in \mathbb{R}^{(T^{'}+1) \times H^{'} \times W^{'} \times D}$. Finally, a sensing matrix reduces the dimension of $\mathbf{p}$ from $D$ to $d$, resulting in the video latent $\mathbf{z}$. The right side of ~\cref{fig:main}(a) depicts the inverse process, where $\mathbf{z}$ is decoded back into the reconstructed video $\mathbf{\hat{x}}$. We set $c_s = 8, c_t = 4, d_1 = 128, d_2 = 384, D = 512 $ and $d \in {4, 16}$. Detailed descriptions  of each component are provided in the following sections.

\subsection{Architecture Details}
\noindent\textbf{Patchify and UnPatchify. }
For the given input $\mathbf{x} \in \mathbb{R}^{(T+1) \times H \times W \times 3}$, we process the first frame $\mathbf{x_0} \in \mathbb{R}^{1 \times H \times W \times 3}$ and following frames $\mathbf{x_{1:T}} \in \mathbb{R}^{T \times H \times W \times 3}$ separately for the joint encoding of videos and images. Specifically, for the continuous frames $\mathbf{x_{1:T}}$, we apply a 3D wavelet transform to map them into LC coefficients (the first subband) and HC coefficients (the remaining subbands concatenated along the channel dimension). The shape of these components are $\frac{T}{2} \times \frac{H}{2} \times \frac{W}{2} \times 3$ for LC and $\frac{T}{2} \times \frac{H}{2} \times \frac{W}{2} \times 21$ for HC.  These features are then split into non-overlapping patches, with a patch size of $2 \times 4 \times 4$. After that, we project the LC and HC patches with two linear layers to obtain their patch embeddings $\mathbf{p_{1:T}^{L}}$ and $\mathbf{p_{1:T}^{H}}$. Similarly, the first frame $\mathbf{x_0}$ undergoes a 2D Haar transform and is divided into $4 \times 4$ patches to obtain the corresponding low-frequency and high-frequency patch vectors $\mathbf{p_0^L}$ and $\mathbf{p_0^H}$. The $\mathbf{p_{0}^{L}}$ and $\mathbf{p_{1:T}^{L}}$ are then concatenated along the temporal dimension as the LC embedding $\mathbf{p^{L}}$, and the same process is applied to the HC embeddings to form $\mathbf{p^{H}}$.   The unpatchify process mirrors the patchify operation. We use the linear layer to project the patch embeddings back to the frequency domain coefficients, followed by resizing and applying the IDWT to reconstruct the RGB visual output $\mathbf{\hat{x}}$. 

Notably, our approach distinguishes from traditional vision transformer patching in three points: 1) we apply the separate patch operation for the first frame and the remaining frames, enabling our model to handle both images and videos. 2) We enrich the representation space by operating in the frequency domain rather than the traditional RGB space. 3) We do not apply normalization to visual patches. We observed it degrades the performance and may result in block artifacts reported in transformer models ~\cite{li2022variable}.

\noindent\textbf{Encoder and Decoder.} 
We build a lightweight \textit{Encoder} network $ \xi $ to further process $\mathbf{p^{L}}$ and $\mathbf{p^{H}}$, enabling feature interaction and transformation to obtain the final feature vector $\mathbf{p}$. Based on the ablation studies of the model architecture (See ~\cref{section:model_analysis}), we adopt a structure that first processes the low- and high-frequency features separately before fusing them. The  $ \xi$ consist of three modules, that is $ \xi_l$ to process $\mathbf{p^{L}}$, $ \xi_h$ to process $\mathbf{p^{H}}$, and $ \xi_f$ for fusing two features. This process can be expressed as:
\begin{equation}
	\mathbf{p} = \xi_f(cat(\xi_l(\mathbf{p^L}), \xi_h(\mathbf{p^H})))
\end{equation}
where $cat$ denotes concatenation along the channel dimension.

We use the Neighborhood-Aware Feedforward with Residual Connection (ResNAF) module as the backbone to build the network $ \xi  = \{ \xi_l, \xi_h, \xi_f\}$, as shown in ~\cref{fig:main}(b). In ResNAF, the 3D input vector first goes through a depthwise convolution with kernel size $3\times3\times3$ to efficiently aggregate local neighboring contexts. The contextual features are then further enriched and transformed through a feedforward layer. Meanwhile, residual connections are incorporated to optimize feature propagation. We construct the $\xi_l, \xi_h, \xi_f $ by stacking 2, 2, and 4 layers of ResNAF, respectively. Through the hierarchical stacking of ResNAF modules, \textit{Encoder} achieves effective information propagation and feature modeling. The \textit{Decoder} $\psi$ follows a symmetric structure to the \textit{Encoder}. 

Notably, we apply causal padding in the temporal dimension for depthwise convolution, ensuring that each frame only interacts with previous frames. This causal property offers several benefits: 1) It improves modeling for images~\cite{yu2024languagemodelbeatsdiffusion}, as they are independent of neighboring frames. 2)  It supports downstream models operating on causal latent spaces~\cite{Li2024AutoregressiveIG}, broadening the applicability of LeanVAE. 3)  It enables efficient temporal tiling inference by caching frames from previous chunks, ensuring continuity in convolutional operations, see ~\cref{sec:caching}. 

\noindent\textbf{Channel Compression Bottleneck. }
This module acts as a latent channel bottleneck, performing both downsampling and upsampling operations. The downsampling part compresses the feature $\mathbf{p}$ from channel dimension $D$ to $d$, obtaining the final video latent $\mathbf{z}$, which is subsequently used by the diffusion model. The upsampling part maps the channel of $\mathbf{z}$ back to $D$, generating the feature $\mathbf{\hat{p}}$ that is fed into the \textit{Decoder} $\psi$.

To implement this, we adopt the classical $\textit{ISTA-Net}^{+}$ framework~\cite{zhang2018ista}, which uses a sensing matrix $\mathbf \Phi \in \mathbb{R}^{d \times D}$ for channel downsampling and a recovery algorithm for upsampling. In the recovery process, the latent $\mathbf{z}$ is first multipied by a learnable matrix $ \tilde{\mathbf \Phi} \in \mathbb{R}^{D \times d } $ to get the initial reconstruction $\mathbf{p_0}$. This is then refined by the recovery network $\mathbf{\Theta}$ to obtain final result $\mathbf{\hat{p}}$. The network $\mathbf{\Theta}$ is a deep unfolding network that simulates an iterative optimization process, with the computation method defined by ~\cref{eq: update r module} and ~\cref{eq: update x module} at each iteration.  The entire computation process is illustrated in ~\cref{fig:main}(b). We employ a two-layer NAF network to construct the forward network \(\mathcal{F}^{(k)}\) and backward network \(\widetilde{\mathcal{F}}^{(k)}\) , with \(\rho^{(k)}\) and \(\theta^{(k)}\) set as learnable scalars. The number of iterations K is set to 2. 

While previous work often employs a simple AE structure with two linear layers for this purpose, we find this approach suboptimal. Instead, we leverage the $\textit{ISTA-Net}^{+}$ framework to achieve effective downsampling and upsampling for the latent channel. Experiments demonstrate its strong effectiveness, leading to significant performance improvements in our model. We provide a straightforward comparison between AE and CS signal recovery algorithms in ~\cref{sec:ae_cs_comparison} for further clarity.

\subsection{Training Objective}
\label{sec:training_objective}

The entire model is then trained end-to-end. Our loss function includes four components, including a L1 reconstruction loss computed in both the RGB and frequency domains, a perceptual loss~\cite{Zhang_Isola_Efros_Shechtman_Wang_2018}) based on VGG features, an adversarial loss computed using a PatchGAN from work ~\cite{CycleGAN2017}, and a KL regularization term~\cite{Kingma_Welling_2013}. The final loss function is formulated as:
\begin{equation}
	\begin{split}
		\mathcal{L} &= \mathcal{L}_{recon} + \lambda_{lpips} \mathcal{L}_{lpips} + \lambda_{adv} \mathcal{L}_{adv} + \lambda_{K\!L}\mathcal{L}_{K\!L}.
	\end{split}
\end{equation}

%% file: sec/4_experiments.tex
\section{Experiments}
 \begin{table*}[ht]
 \centering
 \renewcommand{\arraystretch}{1.25}
 \scalebox{0.80}{
\begin{tabular}{c cc cccc cccc}
 \toprule[1.2pt]
   \multirow{2}{*}{Method}  & \multirow{2}{*}{Param.} & \multirow{2}{*}{Chn.} & \multicolumn{4}{c}{DAVIS} & \multicolumn{4}{c}{TokenBench}\\
   \cmidrule(lr){4-7} \cmidrule(lr){8-11}
   &&&  \textbf{PSNR} $(\uparrow)$ & \textbf{SSIM}$(\uparrow)$ & \textbf{LPIPS} $(\downarrow)$ & \textbf{rFVD} $(\downarrow)$ & \textbf{PSNR} $(\uparrow)$ & \textbf{SSIM} $(\uparrow)$ & \textbf{LPIPS}$(\downarrow)$ & \textbf{rFVD} $(\downarrow)$\\

\midrule
CV-VAE&182M&4&25.75&0.7345&0.1464&598.55 &30.37& 0.8758& 0.0706&265.99\\
Open-Sora VAE&393M &4&\textbf{26.88}&\textbf{0.7741}&0.1582&611.11 &\underline{31.27}&0.8883&0.0774&282.81\\
OD-VAE&239M&4&26.16&0.7589&0.1173&407.20&30.47&0.8791&0.0618&211.93\\
VidTok&157M&4&\underline{26.50}&\underline{0.7707}&\underline{0.1098}&\underline{358.28}&\textbf{31.38}&\textbf{0.8963}&\underline{0.0526}&\underline{178.80}\\
\rowcolor{gray!12}
LeanVAE (Ours)&40M&4&26.04&0.7629&\textbf{0.0899}&\textbf{322.46}&31.12&\underline{0.8957}&\textbf{0.0432}&\textbf{162.55}\\
\midrule
CogVideoX-VAE&206M&16&29.88&0.8741&0.0773&175.57 &34.39&0.9366&0.0331&75.54\\
WF-VAE&316M&16&29.62&0.8668&0.0628&149.27 &35.11&0.9435&0.0222&48.03\\
Cosmos-Tokenizer&101M&16&29.09& 0.8477&0.1412&241.78 &33.13&0.9191&0.0669&109.27\\
VidTok&157M&16&\textbf{31.06}&\textbf{0.8944}&\textbf{0.0436}&\textbf{103.79} &\textbf{36.12}&\textbf{0.9511}&\textbf{0.0166}&\textbf{40.89}\\
\rowcolor{gray!12}
LeanVAE (Ours)&40M&16&\underline{30.15}&\underline{0.8760}&\underline{0.0461}&\underline{119.48} &\underline{35.71}&\underline{0.9506}&\underline{0.0173}&\underline{44.63}\\
\bottomrule[1.2pt]
\end{tabular}}
    \vspace{-0.4em}
 \caption{\textbf{Quantitative comparison on video reconstruction performance.} Chn indicates the dimension of latent channels. All evaluated models are causal and have a video compression ratio of 4 × 8 × 8. The input resolution in DAVIS is 17×256×256, with the encoded latent of the size 5x32x32. The input video in TokenBench are resized with the short size of 256.
 	The highest result is highlighted in \textbf{bold}, and the second highest result is \underline{underlined}.}
    
\label{tb:MainTable} 
 \vspace{-1.5em}
\end {table*}

\subsection{Experimental Setup}
\noindent\textbf{Training Details }\quad

We train the model using Kinetics-600 dataset ~\cite{carreira2018short}. The videos are resized and cropped to resolutions ranging from \(96 \times 128\) to \(352 \times 288\), with their original aspect ratios preserved. For each training step, we randomly sample 17 consecutive frames from a video. First, we train the model for 600k steps without GAN loss, setting $\lambda_{lpips}=4, \lambda_{adv}=0, \lambda_{K\!L}=1e-7 $. The learning rate is warmed up to 5e-5 and decayed to 1e-5 using a cosine scheduler. After that, We continue training for another 100k steps with a learning rate of 1e-5 for the VAE and the discriminator, setting $\lambda_{lpips}=4, \lambda_{adv}=0.2, \lambda_{K\!L}=1e-7 $. We employ the Adam ~\cite{Kingma_Ba_2014} optimizer ($\beta_1$ = 0.5 and $\beta_2$ = 0.9) and train the model with a batch size of 40 on 8 × NVIDIA A40 GPUs.

\noindent\textbf{Evaluation Datasets and Metrics }\quad
Following \cite{agarwal2025cosmos}, we evaluate the reconstruction performance of the video VAEs on two video benchmarks: DAVIS \cite{Pont-Tuset_arXiv_2017} and TokenBench \cite{agarwal2025cosmos}. We employ four evaluation metrics: Structural Similarity Index Measure (SSIM) ~\cite{wang2004image}, Peak Signal-to-Noise Ratio (PSNR)~\cite{Hore_Ziou_2010}, Learned Perceptual Image Patch Similarity (LPIPS) ~\cite{Zhang_Isola_Efros_Shechtman_Wang_2018}, and reconstruction Frechet Video Distance (rFVD) ~\cite{Unterthiner_Steenkiste_Kurach_Marinier_Michalski_Gelly_2019}. To assess performance across different video resolutions, we resize the videos in the evaluation set. Videos in the DAVIS dataset are processed into a square resolution, while videos in TokenBench are resized based on the shorter side, preserving the original aspect ratio. 

To assess our models on video generation, we train the video diffusion model following Latte ~\cite{ma2024latte} for $256\times256$ video generation with 16 frames.\footnote{We actually train the diffusion model on 17 frames and discard the first frame of the generated video to ensure consistency with the baselines during evaluation.} We choose Latte-XL as the denoiser and replace its original VAE (SD VAE) with LeanVAE. The UCF101 ~\cite{Soomro_Zamir_Shah_2012} and SkyTimelapse ~\cite{Xiong_Luo_Ma_Liu_Luo_2017} are used for class-conditional and unconditional video generation respectively. Following the evaluation guidelines in previous video generation works ~\cite{ma2024latte}, we use Frechet Video Distance (FVD) ~\cite{Unterthiner_Steenkiste_Kurach_Marinier_Michalski_Gelly_2019} to measure the quality of generated videos, computing FVD scores based on 2,048 video clips.

\begin{figure*}[ht]
	\centering    \includegraphics[width=0.95\linewidth]{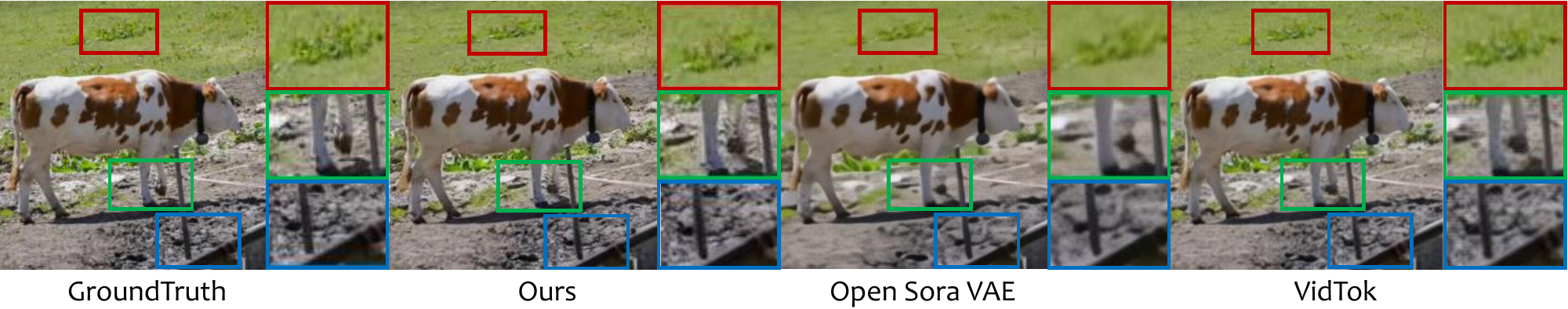}
	\vspace{-0.6em}
	\caption{\textbf{Qualitative comparison between LeanVAE and leading baselines.} Due to space limitations, we present only the model with a latent channel size 4. The reconstruction performance of the leading models with 16 latent channels is notably better, and their visual differences are subtle. More comprehensive visual comparisons are available in the supplementary video.}
	\label{fig:Recon}
	\vspace{-1.5em}
\end{figure*}

\begin{figure*}[ht]
	\centering    \includegraphics[width=\linewidth]{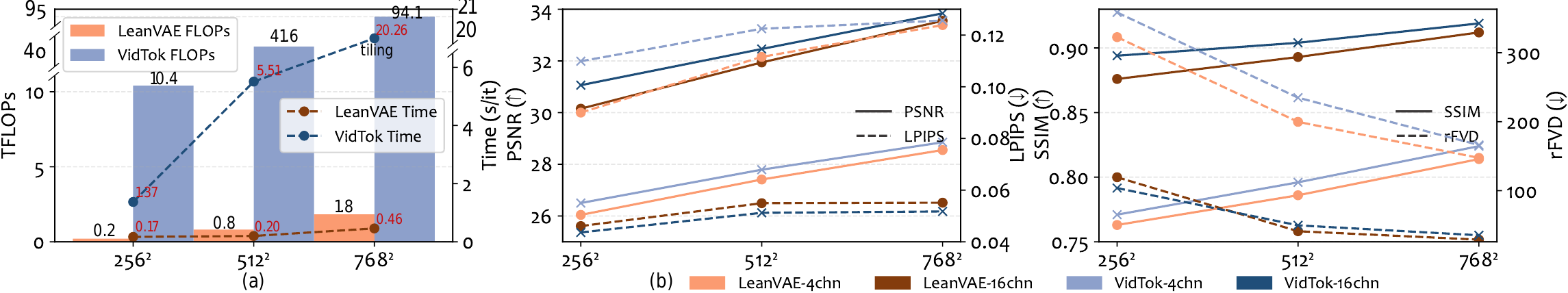}
	
	\vspace{-0.5em}
	\caption{\textbf{Comparison across multiple resolutions}. (a)  Computational cost in terms of TFLOPs(bar plots, labeled in black) and encoding-decoding time (line plots, labeled in red). (b) Reconstruction quality metrics. All evaluations were conducted on 17-frame videos using a single NVIDIA A40 (48GB) GPU.}
	\label{fig:SpeedMemoryTest}
	\vspace{-1.5em}
\end{figure*}

\subsection{Video Reconstruction}

\noindent\textbf{Reconstruction Performance }\quad
To demonstrate the effectiveness of LeanVAE in video reconstruction, we compare it with several state-of-the-art video autoencoders. The baseline models include CV-VAE~\cite{zhao2024cv}, Open-Sora VAE~\cite{opensora}, WF-VAE~\cite{li2024wf}, CogVideoX VAE~\cite{yang2024cogvideox}, VidTok~\cite{tang2024vidtok}, and Cosmos Tokenizer~\cite{agarwal2025cosmos}. VidTok and Cosmos Tokenizer offer both continuous and discrete models; we evaluate only their continuous versions. For all models, we evaluate their latest open-source updates to ensure optimal performance.

The results are presented in ~\cref{tb:MainTable}. LeanVAE achieves highly competitive reconstruction performance with a significantly lean model size. For latent dimensions of 4, our model outperforms all baselines in LPIPS and rFVD metrics, demonstrating superior video quality. It also maintains competing performance in video fidelity, achieving comparable PSNR and SSIM scores with most baselines. When the latent channel is increased to 16, LeanVAE surpasses most models in reconstruction metrics but lags slightly behind VidTok. However, we argue that at 16 latent channels, both LeanVAE and VidTok exhibit strong reconstruction capabilities, and the visual quality gap between them is relatively negligible.
We also provide a qualitative comparison of reconstruction results with competitive baselines in ~\cref{fig:Recon}. As can be seen, our model excels in preserving fine details and capturing motion dynamics. These results underscore the effectiveness of LeanVAE in video reconstruction tasks.

\noindent\textbf{Multi-resolution Evaluation}\quad
\label{sec:Multi-resolution}
We further compare LeanVAE with the best-performing baseline, VidTok, across multiple video resolutions. The evaluation focuses on computational cost and reconstruction performance, with results shown in ~\cref{fig:SpeedMemoryTest}.  

As shown in ~\cref{fig:SpeedMemoryTest}(a), LeanVAE demonstrates remarkable computational efficiency, achieving 50× fewer FLOPs than VidTok and accelerating inference speeds by 8-44× across resolutions. VidTok requires 94.1 TFLOPs for processing a $768^2$ video with 17 frames, demanding impractical GPU memory. Given these,  VidTok and similar models rely on tiling inference\footnote{This involves partitioning the video into chunk groups along temporal ~\cite{chen2024od,li2024wf,tang2024vidtok,agarwal2025cosmos} or spatial dimension~\cite{zhao2024cv,yang2024cogvideox}, followed by sequential inference on each chunk. Caching or overlapping group strategies are employed to maintain convolution continuity, see ~\cref{sec:caching}.}, further slowing down the speed. For instance, VidTok takes 20.26 seconds to process 17 frames of $768^2$ video using 5-frame temporal tiling inference, whereas LeanVAE completes in just 0.46 seconds without chunking the frames. Our tests show that LeanVAE can process a 17-frame 
 1080p video ($3\times1920\times1080$) in 3 seconds, consuming approximately 15GB of GPU memory with FP16 inference.

In addition to its superior inference efficiency, LeanVAE maintains satisfying reconstruction quality, as evidenced in ~\cref{fig:SpeedMemoryTest}(b). At $768^2$ resolution, it achieves near-identical PSNR ($\Delta<$ 0.3dB) to VidTok while delivering better rFVD (18.6 and 6.3 points lower for 4 and 16 channels, respectively).
It is also worth noting that we observed tiling inference may affect the reconstruction quality of VidTok. A well-designed caching strategy can enable lossless temporal titling inference~\cite{li2024wf}, and LeanVAE's structure design supports a simple and efficient temporal caching solution.  Further implementation details are provided in ~\cref{sec:caching}.

Overall, our rigorous evaluation demonstrates that LeanVAE achieves unparalleled efficiency in video reconstruction while maintaining competitive reconstruction quality.

\subsection{Video Generation Evaluation}
\begin{figure*}[ht]
	\centering    \includegraphics[width=0.96\linewidth]{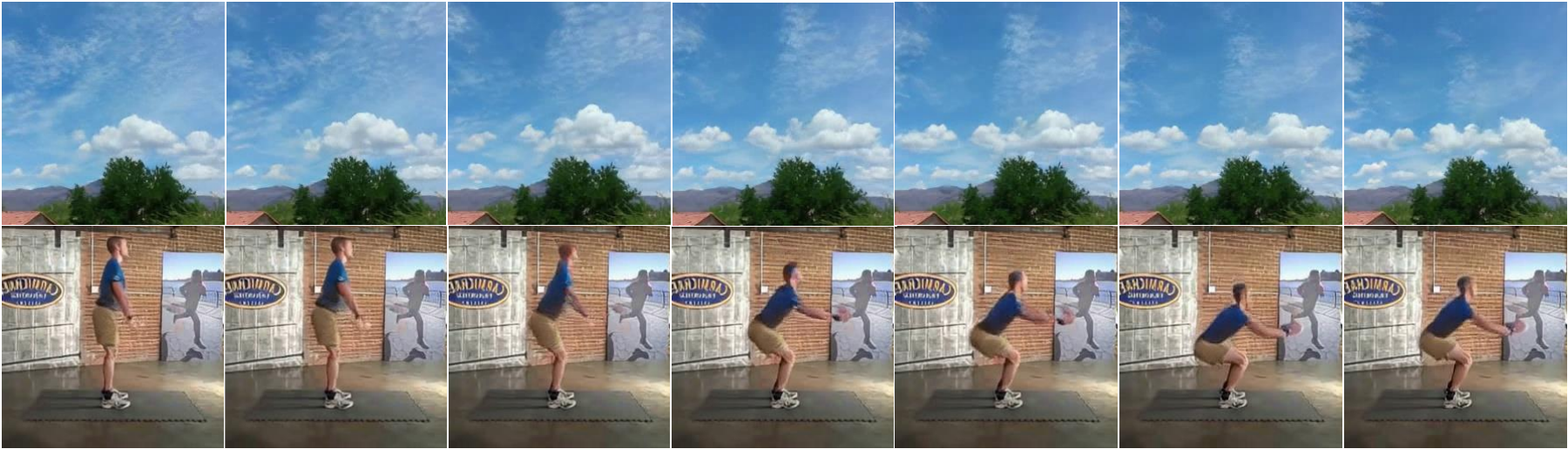}
	\vspace{-0.5em}
	\caption{\textbf{Examples of videos generated by Latte+LeanVAE (chn 4).} The top row is from SkyTimelapse and the bottom is UCF101. More examples are provided in the supplementary video.}
	\label{fig:LatteGenerate}
	\vspace{-1.2em}
\end{figure*}

We evaluate video generation performance in  ~\cref{tb:FVDResults} and present generated examples in ~\cref{fig:LatteGenerate}. As shown, Latte equipped with LeanVAE (using 4 latent channels) achieves a further improvement in generation quality, attaining the best FVD scores for both conditional and unconditional generation. Additionally, due to the 4× temporal compression of inputs and the extra-efficient video encoding, LeanVAE supports a 4× larger batch size and achieves 6.64 samples/sec throughput, leading in a 315\% increase in training speed over the original Latte.

We also notice that increasing the latent channels of LeanVAE to 16, while significantly improving video reconstruction, does not yield corresponding gains in generation performance. The FVD scores actually increased by 45.56 and 10.88, and we observed more distortion in the generated video. This finding aligns with work ~\cite{vavae} that emphasizes the necessity for stricter regularization in high-channel latent spaces. These indicate that training diffusion models in high-channel latent spaces remains an open research area and warrants further investigation.

\begin{figure*}[ht]  
    \centering  
    \begin{minipage}{0.48\textwidth}
        \centering

 \renewcommand{\arraystretch}{1.25}
\scalebox{0.90}{
\begin{tabular}{r|cccc}
\toprule
Method & PSNR $\uparrow$ & SSIM $\uparrow$ & LPIPS $\downarrow$ & rFVD $\downarrow$ \\
\toprule
(a) Variant 1  & 26.04 & 0.755 & 0.143 & 475.17 \\
\rowcolor{gray!12}
Variant 2 & 26.18 & 0.756 & 0.145 & 470.64 \\
Variant 3 & 25.41 & 0.736 & 0.133 & 453.82 \\
\midrule
(b) AE & 25.79 & 0.744 & 0.163 & 535.18 \\
\midrule
(c) w/ norm & 22.91 & 0.703 & 0.158 & 599.38 \\
\bottomrule
\end{tabular}}
\vspace{-0.1em}

        \vspace{-0.7em}
        \captionof{table}{\textbf{Ablation study on different components}. Variant 2 (highlighted in gray) serves as the baseline across all groups, with CS channel compression and without patch normalization.}
        \label{tab:ablation} 
    \end{minipage}%
    \hfill
    \begin{minipage}{0.48\textwidth}
        \centering
        \includegraphics[width=\linewidth]{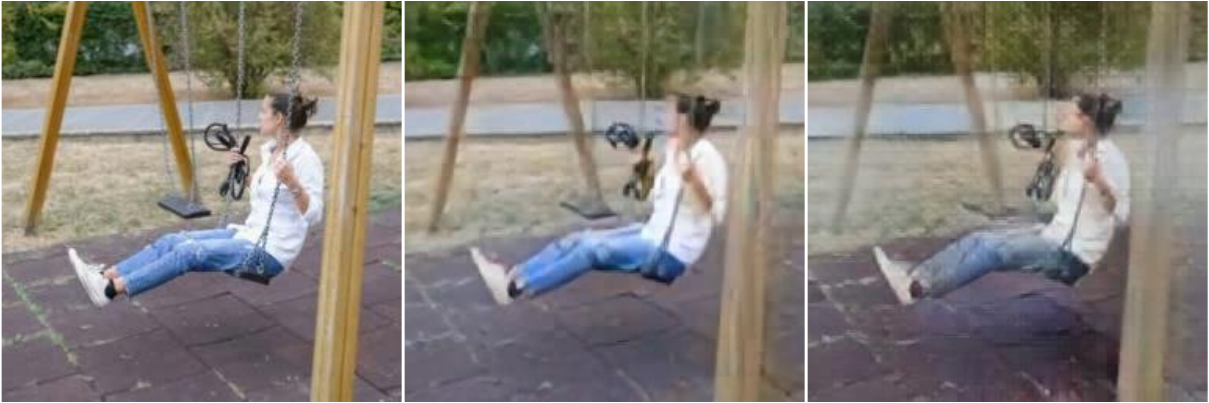}

        \caption{\textbf{Examples of block artifacts in reconstructed video}. Left:ground truth; Middle:reconstruction of model w/o pixel normalization; Right:reconstruction of model w/ pixel normalization.}
        \label{fig:layernorm}
    \end{minipage}
    \vspace{-0.8em}
\end{figure*}

 \begin{table}[h]
        \centering
        \renewcommand{\arraystretch}{1.25}
        \scalebox{0.84}{
        \begin{tabular}{lccc}
            \toprule
            Method & Sample/Sec & SkyTimelapse & UCF101 \\
            \midrule
           VideoGPT~\cite{yan2021videogpt} & - & 222.7 & 2880.6  \\
           StyleGAN-V~\cite{skorokhodov2022stylegan} & - & 79.52 & 1431.0  \\
         LVDM~\cite{he2022latent} & - & 95.20 & 372.0  \\
          Latte~\cite{ma2024latte} (4 chn) & 1.60 & 59.82 & 477.97  \\

            \rowcolor{gray!9}
            Ours-Latte (4 chn) & 6.64  & \textbf{49.59} & \textbf{164.45} \\
            \rowcolor{gray!9}
            Ours-Latte (16 chn) & 6.64 & 95.15 & 175.33  \\
            \bottomrule
        \end{tabular}
        }
        \vspace{-0.7em}
        \caption{\textbf{FVD values of video generation on UCF101 and SkyTimelapse.} FVD for other baselines are referenced from Latte~\cite{ma2024latte} paper. Sample/Sec is measured at the maximum batch size (2 for original Latte and 8 for our Latte) on one A40 (48GB). }
        \label{tb:FVDResults}
    \end{table}

\vspace{-1.5em}
\subsection{Ablation Study}\label{section:model_analysis}
We conduct comprehensive ablation experiments to validate the superiority of the proposed techniques. In all experiments, we set the latent channel dimension to 4 and trained the model for 250k steps. The evaluation is conducted on DAVIS, and the results are summarized in ~\cref{tab:ablation}.

\noindent\textbf{Model Architecture}\quad 
We aim to determine the optimal \textit{Encoder} and \textit{Decoder} configurations for better model performance. To this end, we tested three architectures, including: \textbf{Variant 1}. Processes low-frequency and high-frequency components jointly; \textbf{Variant 2}. Process low-frequency and high-frequency components separately, then merge them; \textbf{Variant 3}. No wavelet transform was applied. The model architectures are plotted in ~\cref{sec:architecture_variants}. All variants were designed to have similar parameter numbers to ensure a fair comparison. As shown in ~\cref{tab:ablation}(a), Variant 2 performs the best fidelity with highest PSNR and SSIM, despite yielding slightly inferior LPIPS and rFVD scores. Given that these perceptual metrics can be greatly optimized through subsequent adversarial training, we choose Variant 2 as the final architecture. The following ablation studies were also conducted based on Variant 2. 

\noindent\textbf{AutoEncoding vs. Compressed Sensing}\quad
We explore the advantages of using compressed sensing (CS) to construct the latent channel bottleneck.  Therefore, we implement the bottleneck using both AE and CS algorithms.  The two approaches have nearly the same parameter counts, except for the additional two scalars $\rho^{(k)}$ and $\theta^{(k)}$ in each iteration of the CS method. As shown in ~\cref{tab:ablation}(b), the CS algorithm $\textit{ISTA-Net}^{+}$ demonstrates significant performance gains, outperforming the AE-based approach by 0.39 dB in PSNR and 0.018 in LPIPS. These results highlight the potential of compressed sensing in this application.

\noindent\textbf{Effect of Patch Normalization}\quad
Block artifacts manifest as grid-like discontinuities in the visual output. We identified the LayerNorm commonly applied to pixel patches in transformers contributes to this issue. We evaluate the model with and without the patch normalization in ~\cref{tab:ablation}(c), showing that normalization leads to a significant decline (3.27 dB in PSNR and 0.013 in LPIPS) in reconstruction metrics. ~\cref{fig:layernorm} visualizes the block artifacts induced by pixel normalization. We hypothesize normalizing pixel patches introduces anisotropy, making it more challenging for the model to maintain consistency across reconstructed patches. We hope this finding could provide valuable insights for related models in low-level vision tasks.

%% file: sec/6_conclusion.tex
\section{Conclusion}
We introduced \textbf{LeanVAE}, a novel and highly efficient Video VAE framework designed to address computational bottlenecks of video compression in Latent Video Diffusion Models. LeanVAE strikes an impressive balance between efficiency and reconstruction quality through the following contributions: (1) a lightweight architecture based on non-overlapping patching and a Neighborhood-Aware Feedforward (NAF) module, (2) the integration of wavelet transforms to enrich input representations, (3) the pioneering application of compressed sensing for channel compression, and (4) an analysis of the model design to identify the optimal architecture. Experiments have verified the boost LeanVAE brings to both video reconstruction and generation.

\noindent\textbf{Limitations and Future Work}\quad Future work includes developing models with higher compression rates, exploring discrete latent space representations, and exploring broader applications of LeanVAE in LVDMs, such as text-to-video generation. These efforts aim to further enhance LeanVAE's capacity and broaden its applicability in the visual generation field.

%% file: sec/X_suppl.tex
\clearpage        
\appendix        
\setcounter{page}{1}  
\onecolumn       
\onecolumn
\begin{center}
    {\Large \textbf{LeanVAE: An Ultra-Efficient Reconstruction VAE for Video Diffusion Models}}\\[1.5em]
    {\Large Appendix}
\end{center}
\vspace{1em}
\section{Overview}
\begin{figure}[ht]
    \centering  
    \includegraphics[width=0.8\linewidth]{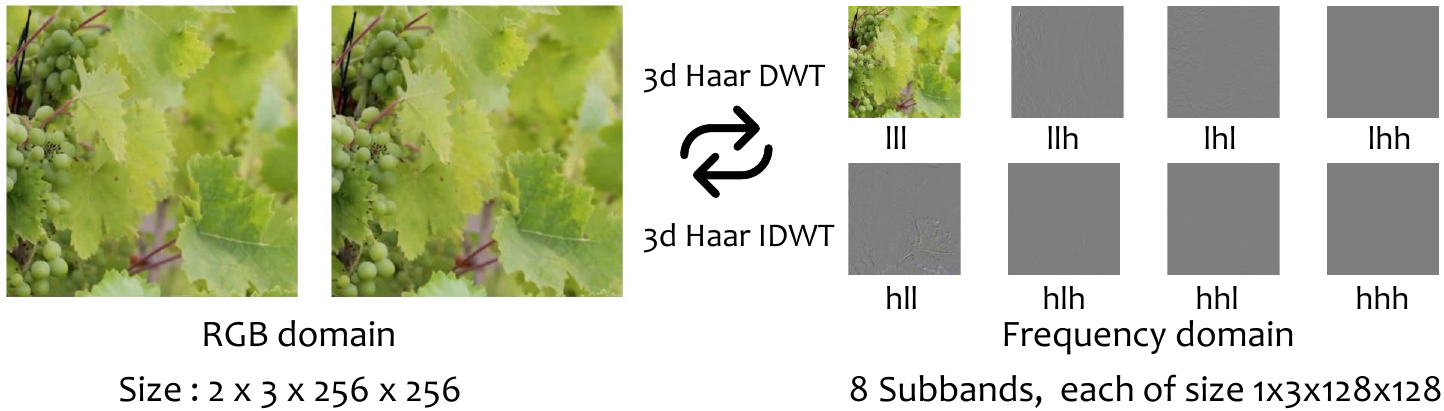} 
    \caption{Visualization of the 3D Haar wavelet transform. }
    \label{fig:wavelet}
    \vspace{-1em} 
\end{figure}

In this appendix, we provide additional details and explanations as follows:
\begin{itemize}
    \item We visualize the transformation of videos into the frequency domain using the Haar Discrete Wavelet Transform (DWT) and the subsequent reconstruction via the Inverse Discrete Wavelet Transform (IDWT) in \cref{fig:wavelet}.
    \item A straightforward comparison between AutoEncoding (AE) and Compressed Sensing (CS) signal recovery algorithms is presented in \cref{sec:ae_cs_comparison}.
    \item We introduce the tiling inference technique utilized in Video VAEs in \cref{sec:caching}, along with its influence on VidTok's reconstruction results in \cref{tab:vidtok_tiling}. Furthermore, we elaborate on the caching strategy adopted by LeanVAE that supports lossless temporal tiling inference.
    \item The architectural details of three LeanVAE variants are plotted in \cref{fig:variants}.
\end{itemize}

\section{Visualization of Haar Wavelet Transform}

\section{Comparison between AE and CS Signal Recovery Algorithms}
\label{sec:ae_cs_comparison}
\begin{figure}[ht]
    \centering    \includegraphics[width=.8\linewidth]{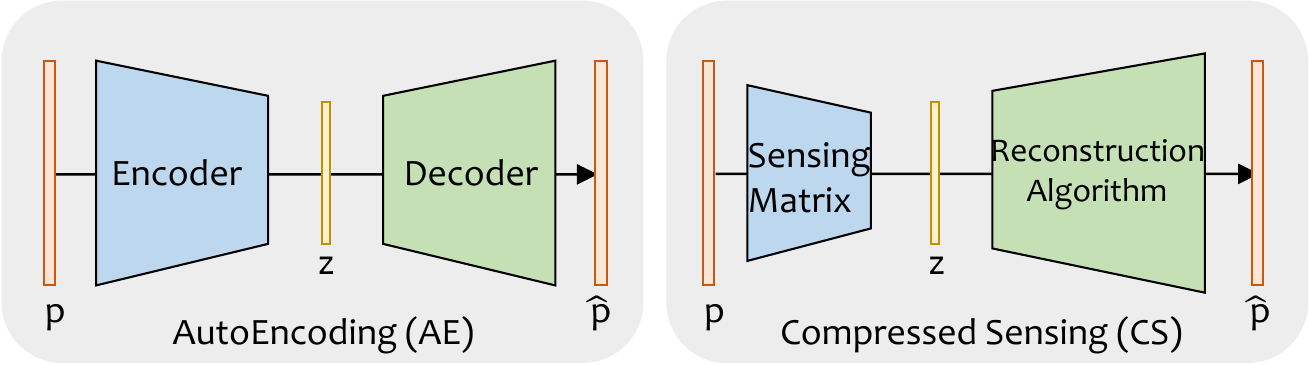}
    \caption{Comparison of AE and CS frameworks.}
    \label{fig:cs}
\end{figure}

AutoEncoding (AE) and Compressed Sensing (CS) are two classical paradigms for data compression and reconstruction. In AE algorithms, data is compressed through an encoder network and subsequently restored by a decoder network. CS methods perform downsampling by multiplying data with a sensing matrix, the signal is then reconstructed by a recovery algorithm. A comparison of these two frameworks is illustrated in ~\cref{fig:cs}.

\section{Tiling Inference}
\label{sec:caching}
\begin{figure}[ht]
    \centering 
    \includegraphics[width=1\linewidth]{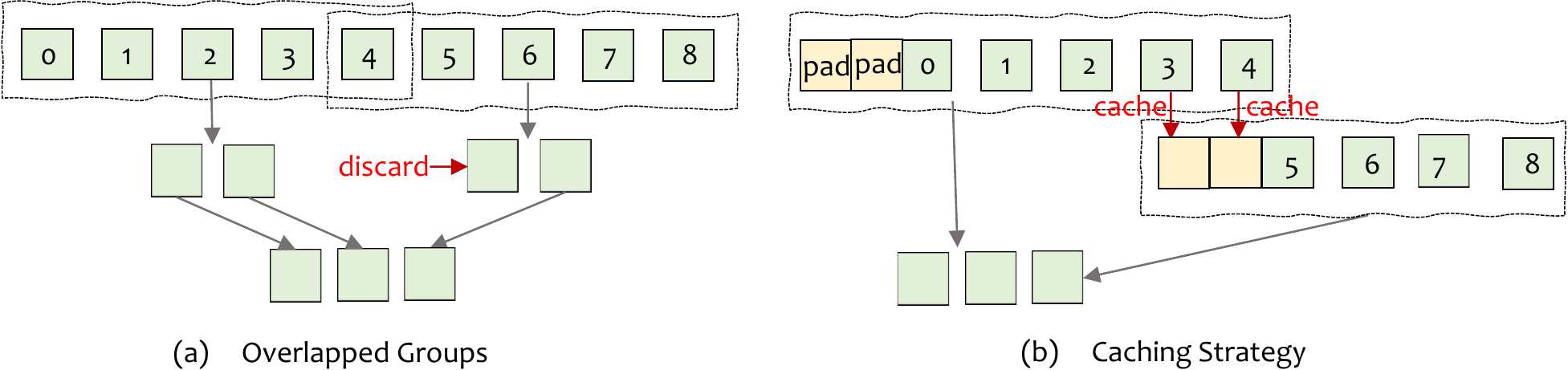} 
    \caption{Two approaches to mitigating discontinuities in temporal tiling inference. Only temporal changes are shown here.  (a) Overlapped group processing, (b) Caching mechanism in convolution modules. The illustration shows chunksize=5, overlap count=1, and cache count=2.}
    \label{fig:caching}
    \vspace{-1em}
\end{figure}
Tiling inference is a widely adopted technique to enable inference on long video sequences or high-resolution frames when constrained by memory limitations. The core idea of tiling inference is to partition the video into smaller chunks, along the temporal \cite{chen2024od,li2024wf,tang2024vidtok,agarwal2025cosmos} or spatial dimension \cite{zhao2024cv,yang2024cogvideox}, with inferenced results subsequently concatenated. However, tiling inference inevitably introduces discontinuities at chunk boundaries due to the lack of contextual information from adjacent chunks during convolution. This discontinuity can lead to visible artifacts or temporal inconsistency in the reconstructed video. 

In this work, we focus on temporal tiling inference, which can mitigate such discontinuities using the properties of causal CNNs. Two common methods are employed to address this issue: (1) Overlapped Groups: As shown in ~\cref{fig:caching}(a), frames are grouped with some overlap between adjacent chunks. (2) Caching Strategy: In convolution, previous inference results are cached as padding for the next chunk, as illustrated in ~\cref{fig:caching}(b). The effectiveness of these approaches depends critically on the chunksize and overlap/cache counts, which influence both the computational efficiency and reconstruction results.

In our multi-resolution experiments (~\cref{sec:Multi-resolution}), as VidTok does not offer caching support, we employ overlapped group methods. At $768^2$ resolution, the maximum feasible chunksize for VidTok is 5 and we set overlap counts to 1. Despite using the overlapped groups method, we observe that VidTok’s reconstruction is still impacted by tiling inference, as shown in ~\cref{tab:vidtok_tiling}. With smaller chunksize , VidTok’s memory usage decreases, but the number of inference steps increases accordingly.

Recent work ~\cite{li2024wf} introduced a model-aware caching strategy for lossless temporal tiling inference. For the LeanVAE model, which consists only of causal depthwise convolutions with a kernel size of 3x3x3 and stride 1x1x1, the lossless caching strategy is straightforward: the last two tokens in previous chunk need to be cached in each convolution structure (the first chunk still uses zero padding).  The detailed computation process can be found in ~\cite{li2024wf}. This approach maintains consistent reconstruction quality across varying chunksizes (~\cref{tab:vidtok_tiling}). Furthermore, as LeanVAE’s memory usage does not vary significantly between 5-frame and 17-frame video inference, the caching operation actually increases memory consumption than full inference.

\begin{table}[!ht]
\centering
\begin{tabular}{lccccc}
\toprule
Model & PSNR $\uparrow$ &  SSIM $\uparrow$ & LPIPS $\downarrow$ & rFVD $\downarrow$ & Mem.(GB) \\
\midrule
VidTok (chunksize=5) & 26.53 & 0.7708 & 0.1086 & 342.44 & 6.5 \\
VidTok (chunksize=9) & 26.48 & 0.7692 & 0.1101 & 351.44 & 8.9\\
VidTok (Full) & 26.50 & 0.7707 & 0.1098 & 358.28  & 13.8\\
\midrule
LeanVAE (chunksize=5) & 26.04 & 0.7629 & 0.0899&  322.56& 2.1\\ 
LeanVAE (chunksize=9) & 26.04 & 0.7629 & 0.0900 & 322.58 & 2.3\\
LeanVAE (Full) & 26.04 & 0.7629 & 0.0899 & 322.46  & 2.3\\
\bottomrule
\end{tabular}
\caption{Performance with temporal tiling inference. Evaluation is conducted on 17-frame 256×256 DAVIS videos. "Full" denotes inference without tiling.} 
\label{tab:vidtok_tiling}
\end{table}

\section{Architectural Variants}
\label{sec:architecture_variants}
\begin{figure}[ht]
    \centering    \includegraphics[width=.8\textwidth]{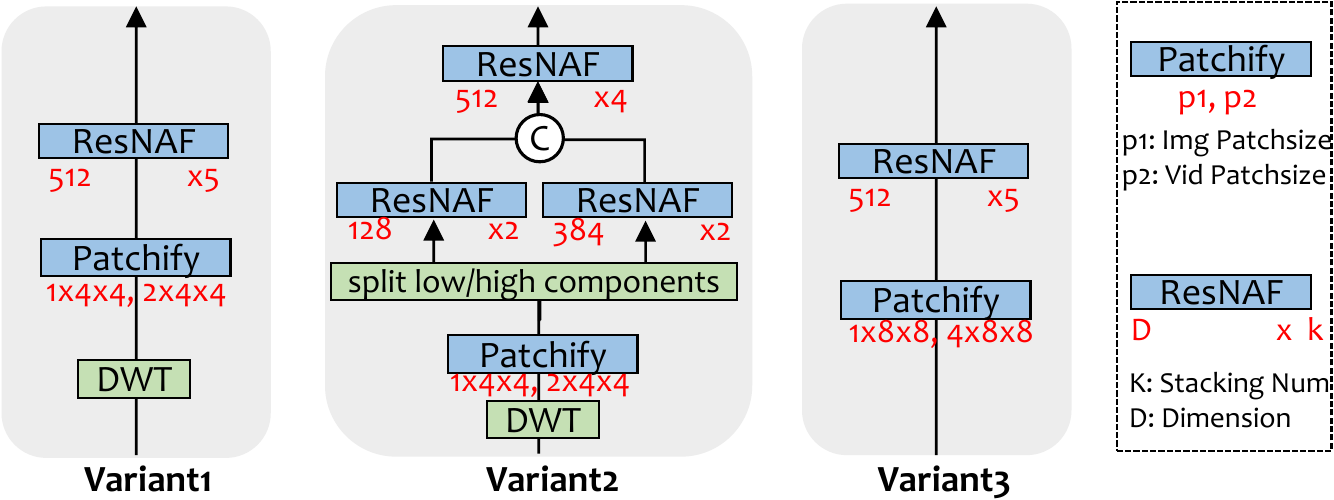}
 
 \caption{Architectural overview of LeanVAE variants (Patching and \textit{Encoder}). The Unpatching and \textit{Decoder} adopt symmetric structures.}
    \label{fig:variants}
    
\end{figure}

We explored three architectural variants of LeanVAE.The detailed network architectures are illustrated in \cref{fig:variants}.
\begin{itemize}
    \item \textbf{Variant 1}: Jointly processes low-frequency and high-frequency components.
    \item \textbf{Variant 2}: Separately processes low-frequency and high-frequency components, followed by merging them.
    \item \textbf{Variant 3}: Directly processes the input without wavelet decomposition.
\end{itemize}